# Using Tree-Decomposable Structures to Approximate Belief Networks

Sumit Sarkar
Department of Quantitative Business Analysis
College of Business Administration
Louisiana State University
Baton Rouge, LA 70803

## Abstract

Tree structures have been shown to provide an efficient framework for propagating beliefs [Pearl,1986]. This paper studies the problem of finding an optimal approximating tree. The star-decomposition scheme for sets of three binary variables [Lazarsfeld,1966; Pearl,1986] is shown to enhance the class of probability distributions that can support tree structures; such structures are called *tree-decomposable* structures. The logarithm scoring rule is found to be an appropriate optimality criterion to evaluate different *tree-decomposable* structures. Characteristics of such structures closest to the actual belief network are identified using the logarithm rule, and greedy and exact techniques are developed to find the optimal approximation.

## 1 INTRODUCTION

Network structures, called *belief networks*, are found to provide an effective framework for probabilistic representation of uncertainty. Unfortunately, schemes to propagate beliefs are of exponential complexity for the general class of belief networks [Cooper,1990]. In real world environments, an expert system must make inferences in a short time. Therefore, an important criterion in representing uncertainty in expert systems is that it allow for efficient manipulation of beliefs. Pearl [1986] has developed a theoretically consistent and computationally efficient belief propagation scheme for a special class of belief networks, namely trees. Pearls' work shows that tree structured representations provide a good framework to represent uncertainties for such environments.

The advantages of using tree structures have been widely recognized. Well-documented implemen-tations such as PROSPECTOR [Duda et al.,1979] make strong independence assumptions in order to use tree structures. However, forcing this assumption where inappropriate leads to encoding probabilities that are inconsistent with the experts' beliefs. Practitioners typically 'adjust' the experts' probability assessments, in order to approximate the true beliefs in the best possible manner. This adjustment of parameters is usually done in an ad-hoc fashion, without considering its implications for the rest of the network. Propagating probabilities in such a network leads to more inaccuracies during the inference process, further compounding the errors.

This research addresses the problem of determining tree representations that approximate the belief network underlying a problem domain. We show that the star-decomposition scheme [Lazarsfeld,1966; Pearl,1986] is applicable for all sets of three dependent binary variables, and can be used to perform belief propagation with no loss of dependency information. This enhances the class of probability distributions that can support tree structures. Classes of probability distributions that support tree structures are identified; structures associated with such distributions are called *tree-decomposable* structures. The problem, then, is one of finding the tree-decomposable representation that is 'closest' to the actual belief network. The logarithm scoring rule [Good,1952] is identified as an appropriate criterion to evaluate approximate representations. The solutions obtained when using this measure are shown to preserve a large number of lower-order marginal probabilities of the actual distribution, and allow for efficient modeling techniques. Finally, greedy and exact techniques are developed to solve for the best representation.

## 2 BELIEF NETWORKS AND BELIEF TREES

Belief networks are directed acyclic graphs in which nodes represent propositions, and arcs signify dependencies between the linked propositions (we use the term *variables* inter-changeably with *propositions*). The belief accorded to different propositions are stated as probabilities (prior or posterior, as the case may be), and the strengths of the dependencies are quantified by conditional probabilities. A collection of propositions with associated dependencies can be conveniently represented using a belief network as shown in Figure 1(a). The nodes A, B, C, D and E denote propositions. Each arc between two nodes represents a dependency across these events, and the direction of the arc indicates an ordering of the events. For instance, in Figure 1(a), nodes A and B are predecessors of C. This indicates that the dependencies between the events A, B and C are represented by storing the conditional probability that event C is true for each realization of the events A and B. The absence of a link between two nodes indicates that the associated events are not directly related. Instead, their dependence is mediated by nodes that lie on the paths connecting them. In probabilistic terms, this means that the two nodes are conditionally independent of each other, given the intermediate nodes on the path between them. In Figure 1(a), the nodes A and D are conditionally independent of each other given realizations for the nodes B and C. A comprehensive discussion of belief networks is provided in Pearl [1988].

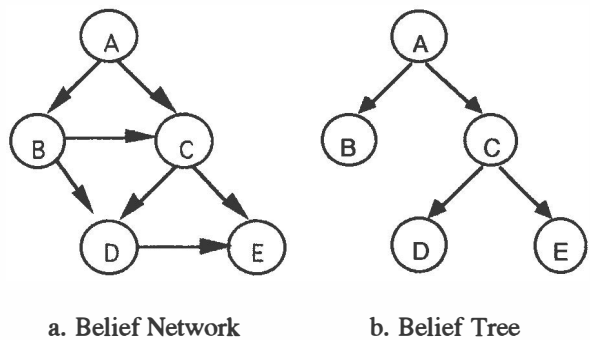

a. Belief Network    b. Belief Tree

Figure 1: Belief Networks and Trees

A belief tree is a special class of belief networks. In tree structures, each node has exactly one parent node, except for the root node which has none. An example of a belief tree is shown in Figure 1(b). In a tree structure, every node is conditionally independent of its parent sub-tree given its immediate parent node. This allows us to represent the tree structure in Figure 1(b) as: $P(A,B,C,D,E) = P(E|C) \cdot P(D|C) \cdot P(C|A) \cdot P(B|A) \cdot P(A)$. A general representation for distributions that support tree structures is:

$$P_T(X_1, X_2, ..., X_n) = P_T(Y_1) \prod_{Y_i \in \underline{X}} P_T(Y_i | F(Y_i));$$

where $Y_1, ..., Y_n$ is some ordering of the variables $X_1, ..., X_n$, and, $F(Y_i)$ refers to the parent node of the variable $Y_i$. $Y_1$ is the node chosen to be the root of the tree and therefore it has no parent.

This simplicity of representation facilitates the modeling process itself, in addition to being computationally efficient. However, this efficiency is obtained by assuming a large number of conditional independences in the network. When such conditions are not met, enforcing a tree structure can require very widespread modification of probability parameters in the belief network.

## 2.1 STAR-DECOMPOSITION

Current implementations usually enforce tree structures by making assumptions regarding conditional independence wherever necessary. We show that it is possible, however, to structure networks as trees with less restrictive assumptions by using auxiliary variables to decompose interdependent variables into conditionally independent ones. The procedure for finding such auxiliary variables is called *star-decomposition*, and it is based upon an analysis performed by Lazarsfeld [1966][1] to discover the existence of latent phenomena on observing manifest data. A set of dependent variables $X_1, ..., X_n$ are said to be star-decomposable if a single auxiliary variable W renders them conditionally independent of each other with respect to the auxiliary variable. The resulting distribution may be represented as:

$$P_S(X_1, ..., X_n, W) = P_S(W) \prod_i P_S(X_i | W).$$

The procedure presented by Lazarsfeld uses the joint distribution of the variables of interest (hereafter called *observable* variables) to determine parameters that specify the star structure. These parameters are: the probabilities associated with the auxiliary variable, and, the conditional probabilities of the observable variables with respect to each outcome of the auxiliary variable. Consider the case where three binary variables are to be star-decomposed. The joint distribution for the three variables can be completely specified in terms of the following seven joint-occurrence probabilities:

$$p_i = P(X_i) \qquad \forall \ i = 1,2,3$$
$$p_{ij} = P(X_i, X_j) \qquad \forall \ i,j = 1,2,3 \text{ and } i \neq j$$
$$p_{ijk} = P(X_i, X_j, X_k) \qquad i,j,k = 1,2,3$$

The probability of a binary auxiliary variable and the conditional probabilities of the observable variables with respect to the auxiliary variable are represented by the following terms:

- $P_S(W)$
- $P_S(X_i | W) \qquad \forall \ i = 1,2,3$
- $P_S(X_i | \neg W) \qquad \forall \ i = 1,2,3$

Since $X_1$, $X_2$ and $X_3$ have to be conditionally independent for all realizations of W, the joint occurrence probabilities can be represented in terms of the conditional probabilities as follows:

$$P(X_i) = P_S(X_i | W) \cdot P_S(W) + P_S(X_i | \neg W) \cdot P_S(\neg W)$$
$$\forall \ i = 1,2,3 \qquad (i)$$

---

[1] Subsequently discussed in Pearl [1986].

$P(X_i, X_j) = P_S(X_i \mid W) \cdot P_S(X_j \mid W) \cdot P_S(W) +$
$\qquad P_S(X_i \mid \neg W) \cdot P_S(X_j \mid \neg W) \cdot P_S(\neg W)$

$\qquad \forall\ i,j = 1,2,3 \text{ and } i \neq j \qquad$ (ii)

$P(X_i, X_j, X_k) = P_S(X_i \mid W) \cdot P_S(X_j \mid W) \cdot P_S(X_k \mid W) \cdot P_S(W) +$
$\qquad P_S(X_i \mid \neg W) \cdot P_S(X_j \mid \neg W) \cdot P_S(X_k \mid \neg W) \cdot P_S(\neg W)$
$\qquad\qquad\qquad\qquad\qquad\qquad\qquad$ (iii)

The above expressions translate to seven non-linear equations with seven unknown variables (the variable $P_S(W)$, and, $P_S(X_i \mid W)$ and $P_S(X_i \mid \neg W)$ $\forall$ i=1,2,3). Lazarsfeld's procedure is used to solve the above system of equations. Figure 2 illustrates the star structure obtained for the three binary variables $X_1$, $X_2$ and $X_3$. Although the three variables are interrelated, yet with the introduction of the auxiliary variable W, such inter dependencies are replaced by their respective dependencies on the auxiliary variable. No information is lost so long as the marginal distribution for $X_1, X_2$ and $X_3$ in the distribution $P_S(\cdot)$ is the same as that for the variables in the original distribution $P(\cdot)$.

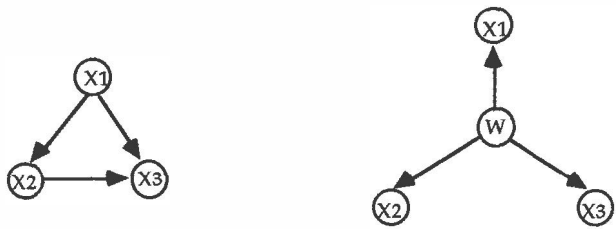

Figure 2: Star-decomposition for Three Binary Variables

It is easy to see that star structures are a special form of tree structures. Further, any of the observable variables may be made the root of the star (tree) structure by performing suitable arc reversals; for instance, in Figure 2 the arc from W to $X_1$ can be reversed by using Bayes rule to evaluate the conditional probabilities $P_S(W \mid X_1)$ and $P_S(W \mid \neg X_i)$, and storing these parameters instead.

The solution procedure discussed by Lazarsfeld works for all sets of three mutually dependent binary variables. However, the procedure does not guarantee that solutions satisfy probability axioms, i.e. they lie in the interval [0,1]. Since the variables of interest are all defined as probabilities, solutions in [0,1] can be easily interpreted in a probabilistic sense. Pearl [1986] has derived necessary and sufficient conditions for the solutions to satisfy probability axioms. An intuitive interpretation of these conditions is that the three variables must be positively correlated, and the correlation between any two variables must be stronger than that induced by their dependencies on the third variable. These conditions place stringent requirements on the random variables for existence of star-decomposability and are unlikely to be satisfied in a lot of cases. Conditions are needed under which most, if not all, sets of three mutually dependent variables can be represented using tree-structures. We state two important results about star-decomposition solutions, proofs of which appear in [Sarkar,1991]:

*Proposition 1*: The star-decomposition procedure for three binary variables leads to solutions that have a unique interpretation. •

*Proposition 2*: Star-decomposition solutions, whether they satisfy probability axioms or not, can be used to consistently update the probabilities associated with observable variables. •

The star-decomposition procedure involves solving a quadratic equation, and therefore has two roots. We are able to show that the second root corresponds to a solution where the resulting auxiliary variable is equivalent to the negation of the auxiliary variable obtained from the first root. *Proposition 2* is illustrated with the help of an example. Consider three variables $X_1$, $X_2$ and $X_3$ with the following joint occurrence probabilities:

$P(X_1) = 0.7 \qquad P(X_2) = 0.56 \qquad P(X_3) = 0.41$

$P(X_1, X_2) = 0.428 \quad P(X_1, X_3) = 0.278 \quad P(X_2, X_3) = 0.226$

$P(X_1, X_2, X_3) = 0.1708$

Solving for the parameters that describe the star-decomposed structure, we get:

$P_S(X_1 \mid W) = 0.8 \qquad P_S(X_1 \mid \neg W) = -0.2$

$P_S(X_2 \mid W) = 0.6 \qquad P_S(X_2 \mid \neg W) = 0.2$

$P_S(X_3 \mid W) = 0.4 \qquad P_S(X_3 \mid \neg W) = 0.5$

$P_S(W) = 0.9$

The solution may be easily verified from equations *(i)*, *(ii)* and *(iii)* (the other solution that is obtained is one where $P_S(W) = 0.1$, and the conditional probabilities given W and its negation, respectively, are interchanged). Although $P_S(X_1 \mid \neg W)$ is not in [0,1], and thus cannot be interpreted as a probability measure, it can be used to update the probability of any one of the observable events $X_1$, $X_2$ or $X_3$, given some other event is observed to be true or false. For instance, if event $X_2$ is known to be true, then we can update the probability of event $X_1$ using the parameters associated with the star structure:

$P_S(X_1 \mid X_2) = P_S(X_1 \mid W) \times P_S(W \mid X_2) +$
$\qquad\qquad P_S(X_1 \mid \neg W) \times P_S(\neg W \mid X_2);$

where

$P_S(W \mid X_2) = \dfrac{P_S(X_2 \mid W) \times P_S(W)}{P_S(X_2 \mid W) \times P_S(W) + P_S(X_2 \mid \neg W) \times P_S(\neg W)}$

$\qquad\qquad = \dfrac{54}{56};$ and

$P_S(\neg W \mid X_2) = \dfrac{2}{56}$.

Thus $P_S(X_1|X_2) = \frac{0.428}{0.56}$, which is identical to the value of $P(X_1|X_2)$ evaluated using the original joint occurrence probabilities. The result follows from the fact that the star-decomposition solutions satisfy the set of equations represented by *(i)*, *(ii)* and *(iii)*, and since $P(X_1|X_2) = \frac{P(X1,X2)}{P(X2)}$, the values of the numerator and the denominator are preserved when using the star-decomposition solutions. This result will hold when evaluating the posterior probability of *any* of the three variables based on observing one or both of the other variables.

The star-decomposition procedure is an elegant way to decompose belief network components comprising of three binary variables into tree structures. Unfortunately, when the number of variables involved exceeds three, the conditions for star-decomposability are very restrictive, and unlikely to be met in practice. For a star structure to exist with 4 observable variables, we have to solve a system of 15 equations with only 9 independent parameters which is not feasible in general.

## 2.2 USING STAR-DECOMPOSITION TO OBTAIN TREE STRUCTURES

While the star-decomposition procedure cannot provide exact tree representations for arbitrarily large networks of inter-connected events, it can be used to reduce the assumptions of conditional independence that are made. Figure 3 illustrates how using star-decomposition helps preserve a large number of dependency relationships while using tree structures.

For the example network in Figure 3(a), *simple* tree representations will not preserve many of the dependencies in the actual network. For instance, the direct dependence between the variables B and C are not captured in the tree representation shown in Figure 3(b); instead the tree structure imposes conditional independence of the variables B and C with respect to the variable A. We note that the tree structure shown is one of many possible tree structures that may be used to represent the network.

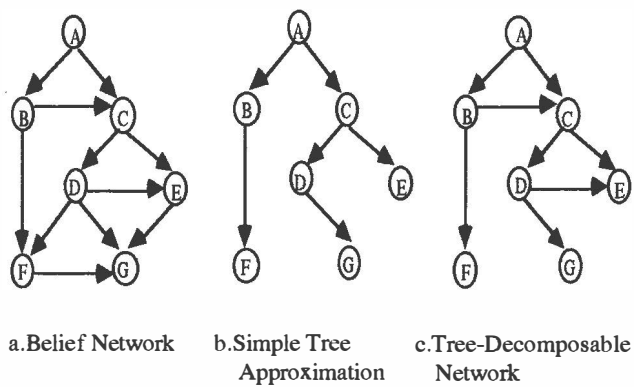

a. Belief Network    b. Simple Tree Approximation    c. Tree-Decomposable Network

Figure 3: A Belief Network and Some Feasible Tree Representations

Tree representations that are obtained by star-decomposing triplets of variables can preserve many more dependencies inherent in the actual belief network. A feasible representation is shown in Figure 3(c). While the structure shown is not a tree, it can be transformed into a probabilistically equivalent tree by star-decomposing the two inter-dependent triplets of variables (A,B,C) and (C,D,E) that appear in the structure. The resulting structure preserves the joint distribution across these two sets of triplets, and therefore reduces assumptions of conditional independence that need to be made to obtain trees. We use the term *tree-decomposable networks* to classify such structures. We note that our use of the term *tree-decomposable* is somewhat different from Pearl's [1986] use of the same term. We postpone a formal definition of this term until the next section, where we also identify the differences.

## 3 OPTIMAL TREE-DECOMPOSABLE NETWORKS

Real-world applications are often not amenable to exact tree representation, even with the help of star-decomposition. Therefore, in order to take advantage of the efficient belief propagation features of tree structures, we must approximate the distribution underlying a problem domain. The use of star-decomposition allows us to consider both tree-decomposable structures and simple tree structures (representations that do not use star-decomposition) as approximate representations. We formulate this problem as one of determining the probability distribution $P_T(\cdot)$ that can support a tree structure (either directly, or with the help of star-decomposition), and that is *closest* to the actual distribution underlying the problem domain $P(\cdot)$ in terms of some measure of closeness $M(P,P_T)$.

### 3.1 DISTRIBUTIONS THAT SUPPORT TREE STRUCTURES

Probability distributions that support simple tree structures can be represented as the product of conditional probability terms that have one conditioned and one conditioning event (as discussed in Section 2.1); hence, they are called *second-order* product distributions. By virtue of star-decomposition, a wider class of probability distributions can now be used to support tree structures. For instance, the tree-decomposable structure in Figure 3(c) can be represented as:

$P_T(A,B,C,D,E,F,G,H) =$

   $P_T(A) \, P_T(BC|A) \, P_T(DE|C) \, P_T(F|B) \, P_T(G|D)$

Thus, probability distributions associated with tree-decomposable networks can include *component distributions* that include two conditioned variables and one conditioning variable. We call them *third-order* product distributions. Therefore, tree-decomposable networks are those networks which support third-order product distributions as defined above. We note that third-order representations typically include both second-order and third-order components. Joint distributions for third-order terms are transformed into tree structures by incorporating auxiliary variables.

Our definition of a tree-decomposable structure is different from Pearl's [1986]. His definition is semantically accurate in that he includes all possible distributions that may be represented as tree structures with the help of auxiliary variables. This includes third- as well as higher order distributions. On the other hand, Pearl considers only those instances of third-order distributions where star-decomposition results in parameters that satisfy probability axioms; our definition includes all instances of third-order distributions. For higher order distributions, Pearls definition includes those instances for which an underlying tree-structured representation is *known to exist*. As he points out, conditions for the existence of such structures are quite restrictive. A more accurate definition for the structures we consider in this paper would be *third-order tree-decomposable networks*; however, for the sake of brevity, we drop the qualifier *third-order* from our definition.

### 3.2 EVALUATING APPROXIMATE DISTRIBUTIONS

A measure is needed to compare different approximations with the given probability distribution. The choice of an appropriate criterion is very important since the 'best' approximation will depend on the criterion chosen, and may be different when different measures are used. We discuss some fundamental desirable properties for an appropriate measure, and identify the logarithm measure as one that satisfies those requirements.

Approximate probability distributions have been analyzed in judging subjective probability assessments made by experts (e.g. weather forecasts by meteorologists). A reasonable measure is a function of the assessed probability distribution and subsequent observation of the actual realization. The term *scoring rule* is used for such measures [Stael von Holstein,1970]. Scoring rules are designed to encourage assessors to provide their true ('honest') estimates, and to evaluate different probability assessments. If Y is an uncertain quantity represented by a discrete probability distribution $\underline{p} = (p_1,..., p_n)$, and the distribution $\underline{r} = (r_1,..., r_n)$ is the assessors stated belief, then the assessor receives a score $S_k(\underline{r})$ when the $k^{th}$ event occurs. The expected score is $S(\underline{r},\underline{p}) = \sum p_k S_k(\underline{r})$. A scoring rule is *proper* if $S(\underline{p},\underline{p}) \geq S(\underline{r},\underline{p})$; therefore, assessments other than $\underline{p}$ cannot get a higher expected score than $\underline{p}$ itself. Attempts have been made to characterize the functional form of proper scoring rules [McCarthy,1956; Marschak,1959; Shuford et al.,1966]. Among all the feasible proper scoring rules, three have received particular attention in the literature. They are:

- Quadratic scoring rule [Brier, 1950], defined as:
  $S(\underline{p},\underline{r}) = -\sum (p_k - r_k)^2$.
- Logarithmic scoring rule [Good, 1952], defined as:
  $S(\underline{p},\underline{r}) = \sum p_k \log r_k$.
- Spherical scoring rule [Roby, 1965], defined as:
  $S(\underline{p},\underline{r}) = \dfrac{\sum p_k r_k}{(\sum r_i^2)^{0.5}}$.

While proper scoring rules are desirable, other factors are also important. Stael von Holstein [1970] lists three principles to help choose scoring rules for evaluating different assessments. These are: (i) Relevance; (ii) Invariance; and (iii) Strong discriminability. The *relevance* principle states that the score $S(\underline{p},\underline{r})$ should depend only on the probability assigned to the event that is actually realized. *Invariance* means that the scoring rule should be independent of the permutation of different events. If events are re-ordered along with their respective probabilities and assessments, the score should not change. The *strong discriminability* principle states that the composite score for compound events that are more likely, if they are realized, should be higher than for less likely compound events (as is the case for simple events). It has been shown that any measure that satisfies the properties of *relevance*, *invariance* and *strong discriminability* must be a linear transformation of the logarithm measure [Sarkar, Mendelson and Storey,1992]. Hence, this measure is chosen as the appropriate *closeness* measure to evaluate different approximate representations for a belief network.

It is important to note that the I-Divergence measure [Kullback, 1959] is a linear transformation of the logarithm scoring rule, and minimizing the I-Divergence measure is equivalent to maximizing the logarithm scoring rule. This equivalence has been observed in many different contexts by various authors (for instance [Good,1952; Savage,1971; Dalkey,1992; etc.]).

### 3.3 SOLUTION CHARACTERISTICS OBTAINED USING THE LOGARITHM RULE

In order to find the best tree representation for a belief network, we must find the second- or third-order product distribution $P_T(\cdot)$ that is closest to the original distribution $P(\cdot)$ with respect to the logarithm rule. This formulation has two kinds of unknown parameters. First, the specific topology of the best approximate tree is not known; that is, the specific combination of variables in each third- or lower-order term is not known. The number of feasible topologies increases exponentially with the number of variables. Second, given the topology, the probabilities for conditional distributions that appear in the best approximation have to be determined. The constraints for the resulting optimization problem are not linear; hence, for a given topology, the second problem becomes one of nonlinear optimization. Therefore, in the worst case, finding the best solution could potentially require solving non-linear optimization problems for a very large number of feasible topologies, and then choosing the best solution among them. We state some theoretical results that are of great importance in solving the optimization problem (proofs in [Sarkar,1991]).

*Proposition 3*: The best third-order product approximation is at least as good as the best second-order approximation, and is strictly superior if there are no conditional independences that exist in the actual probability distribution underlying the problem domain.  •

*Proposition 4*: The best third-order product approximation obtained for a given probability distribution is one that preserves the joint probability distribution for each component of the product approximation.  •

The result of *Proposition 3* is to be expected, since third-order product distributions preserve more dependencies than second-order product distributions. *Proposition 4* states that if the best third-order product approximation for the belief network in Figure 3(a) is as shown in Figure 3(c), then we must have:

- $P_T(A) = P(A)$
- $P_T(BC|A) = P(BC|A)$
- $P_T(DE|C) = P(DE|C)$
- $P_T(F|B) = P(F|B)$
- $P_T(G|D) = P(G|D)$

This result has some very important implications for determining the best tree-decomposable representation. First, it says that once the topology for the best representation is known, the probability distribution for the variables in each component of the tree-decomposable representation can be obtained by finding the appropriate marginal joint distribution for those variables in the actual distribution. The distribution for the complete product approximation is obtained by using the appropriate product form. Thus, the problem of finding the best approximate tree representation is reduced to one of finding the topology of the tree that supports the best representation. Second, this is intuitively a very appealing result, and strongly reinforces Pearls' observation [1986] that humans perform very well with knowledge that is defined over low-order marginal and conditional probabilities, instead of the entries in large joint distribution tables. Since such lower-order distributions are obtained more easily and with greater accuracy, preserving these distributions in approximate representations is very desirable. The result also provides validation for the often used technique of eliciting probability distributions over a large number of variables by first obtaining smaller component distributions, and then combining them to construct the larger joint distributions.

### 3.4 DETERMINING THE BEST TOPOLOGY

Unfortunately, when the number of variables is large, the number of feasible topologies is very high, and, therefore, finding the best topology is potentially a combinatorially hard problem. Furthermore, evaluating the logarithm score for a complete joint distribution is itself computationally intense. We wish to eliminate dominated solutions using local characteristics of different topologies wherever possible. We invoke the well-known definition of *mutual information* across sets of variables [Kullback, 1959], and show that it is useful in comparing different topologies in an efficient manner. The mutual information between two variables A and B is defined as follows:

$$I(A;B) = \sum_{A,B} P(A,B) \log \frac{P(A,B)}{P(A)P(B)}$$

Similarly, the mutual information across three variables A, B and C is given by:

$$I(A;B;C) = \sum_{A,B,C} P(A,B,C) \log \frac{P(A,B,C)}{P(A)P(B)P(C)}$$

We are able to characterize the best third-order solution using mutual information terms in the following manner (proof in [Sarkar,1991]).

*Proposition 5* : The best solution to the approximation problem is one that maximizes the sum of the mutual information terms associated with components of the third-order distribution.  •

The mutual information terms (called *weights*) for second- and third-order components can be evaluated easily, as compared to finding the logarithm score for the entire distribution. As a result, different structures may be easily compared using the *weights* associated with the components of the overall distribution. In order to find the best representation, we need to determine the representation that maximizes the sum of weights associated with it. A sub-optimal greedy algorithm and an exact branch and bound algorithm have been developed. Since the optimal solution may consist of both second- and third-order component distributions, the weights for all possible combinations of pairs and triplets of variables are determined. The weights are sorted in descending order, and the greedy solution is obtained in one pass of these weights as follows. First, the highest weight term is included in the solution. The next highest weight term is included if it does not violate the requirements for a third order distribution. This is easily enforced by checking that the weight under consideration does not involve more that one of the observable variables that have already been included in the solution. This process is repeated until all variables are included in the network. Unfortunately, the greedy method does not guarantee optimal solutions [Sarkar,1991]. The exact solution procedure uses a branch and bound algorithm to identify the best tree-decomposable structure. It generates partial solutions using a best-first approach, and uses the greedy method to obtain lower bounds for each partial solution. The highest lower bound among all generated partial solutions is the *current* best solution to the problem. An upper bound for each partial solution is obtained by relaxing the acyclic requirement for the tree-decomposable network. A newly generated partial solution is eliminated if it's upper bound is dominated by the *current* best solution. This process is repeated until all partial solutions have been fathomed. The technique has been tested for problems involving up to 9 variables. The average number of structures evaluated before obtaining the best solution is around 300 of the approximately $10^{10}$ possible structures for 9 variable problems[2] [Sarkar, 1991]. Furthermore, the greedy procedure is found to converge to the best solution very quickly (under 10 iterations on average), although verifying that a solution is optimal still requires the branch and bound approach. This indicates that we are reasonably likely to obtain optimal solutions even if we terminate the search after a suitably high number of iterations for large problem instances.

## 4 CONCLUSIONS AND FUTURE RESEARCH

Some related issues are currently being addressed. A cost/benefit analysis for using tree-decomposable structures is being performed. The costs of using this technique include determining the best tree structure, and, the increase in time to propagate beliefs due to the addition of auxiliary nodes. The effort in obtaining the structure is not a recurring cost and is therefore not significant. The number of auxiliary variables that may be added is shown to be at most $n/2$ [Sarkar,91]. Since the complexity of belief propagation in trees is polynomial in the number of nodes [Pearl,1986], this increase in the number of nodes is not a severe bottleneck. Currently, work is in progress to compare the performance of tree-decomposable structures with simple tree structures. We assign costs to incorrect decisions made when using approximate representations, and evaluate the expected losses incurred when using approximate structures. Our preliminary results indicate that tree-decomposable structures significantly outperform simple trees, and will often justify the added effort in obtaining and using such structures.

### Acknowledgements

I would like to thank Professor Haim Mendelson from Stanford University, and Professors Henry Kyburg and Veda C. Storey from the University of Rochester, for their comments and suggestions at various stages of this research. I would also like to thank the anonymous referees for their helpful comments.

---

[2] The branch and bound algorithm was implemented and run on a Macintosh SE/30 with 5MB of RAM, and took less than five minutes on average to find the best solution.